\documentclass[conference]{IEEEtran}
\IEEEoverridecommandlockouts

\usepackage{cite}
\usepackage{amsmath,amssymb,amsfonts}
\usepackage{graphicx}
\usepackage{textcomp}
\usepackage{xcolor}
\definecolor{RoyalBlue}{RGB}{65,105,225}

\usepackage{tikz}
\usepackage{standalone}
\usepackage{multirow}
\usepackage{booktabs}
\usepackage[inline]{enumitem}
\usepackage{siunitx}
\usepackage{hyperref}
\usepackage{pgf}

\usepackage{tcolorbox}
\usepackage{algorithm}
\usepackage{algorithmic}
\usepackage{caption}
\usepackage{academicons}
\usepackage{fontawesome5}

\usepackage{amsmath,amsfonts,bm}

\def\eqref#1{equation~\ref{#1}}

\def\1{\bm{1}}

\def\ve{{\bm{e}}}

\def\vv{{\bm{v}}}

\DeclareMathAlphabet{\mathsfit}{\encodingdefault}{\sfdefault}{m}{sl}
\SetMathAlphabet{\mathsfit}{bold}{\encodingdefault}{\sfdefault}{bx}{n}

\def\gE{{\mathcal{E}}}

\def\gG{{\mathcal{G}}}

\def\gV{{\mathcal{V}}}

\hypersetup{
    colorlinks=true,
    allcolors=RoyalBlue
}

\def\BibTeX{{\rm B\kern-.05em{\sc i\kern-.025em b}\kern-.08em
    T\kern-.1667em\lower.7ex\hbox{E}\kern-.125emX}}

\title{Predicting Channel Closures in the Lightning Network with Machine Learning}

\author{
\IEEEauthorblockN{
Simone Antonelli\textsuperscript{1,*},
Vincent Davis\textsuperscript{2},
Harrison Rush\textsuperscript{2},
Anthony Potdevin\textsuperscript{2},\\
Jesse Shrader\textsuperscript{2},
Vikash Singh\textsuperscript{3},
Emanuele Rossi\textsuperscript{2,4}
}
\vspace{0.3em}
\IEEEauthorblockA{
\textsuperscript{1}CISPA Helmholtz Center for Information Security \quad
\textsuperscript{2}Amboss Technologies\\
\textsuperscript{3}Stillmark \quad
\textsuperscript{4}Sapienza University of Rome\\
\vspace{0.3em}
\texttt{simone.antonelli@cispa.de} \quad \texttt{\{v,harrison,ap,j\}@amboss.tech}\\
\texttt{vikash@stillmark.com} \quad \texttt{emanuele.rossi1909@gmail.com}
}
\thanks{\textsuperscript{*}Work done as an intern at Amboss Technologies.}
\thanks{Accepted for publication at the 2026 8th International Conference on Blockchain Computing and Applications (IEEE BCCA 2026). \copyright~2026 IEEE. Personal use of this material is permitted. Permission from IEEE must be obtained for all other uses, in any current or future media, including reprinting/republishing this material for advertising or promotional purposes, creating new collective works, for resale or redistribution to servers or lists, or reuse of any copyrighted component of this work in other works.}
}

\begin{document}

\maketitle

\begin{abstract}
The Lightning Network (LN) is a second-layer protocol for Bitcoin designed to enable fast and cost-efficient off-chain transactions. Channels in the LN can be closed either by mutual agreement or unilaterally through a \emph{forced closure}, which locks the involved capital for an extended period and degrades network reliability. In this paper, we study the problem of predicting channel closure types from publicly available gossip data, framing it as a temporal link classification task over the evolving channel graph. We construct a dataset spanning over two years of LN activity and benchmark a range of machine learning approaches, from MLPs to temporal graph neural networks and spectral encodings. Our experiments reveal that the dominant predictive signals are temporal and behavioural, namely how recently each endpoint was active and the per-node history of past closures, while the surrounding network topology provides no additional benefit. We find that a simple MLP operating on edge-level features, node-level event counts, and temporal patterns outperforms all graph-based approaches, and discuss how the inherent privacy of the LN, where critical information such as channel balances and payment flows remains hidden, fundamentally limits the predictability of closures from gossip data alone. We publicly release the dataset and code at \href{https://github.com/AmbossTech/ln-channel-closure-prediction}{\faGithub\ \texttt{AmbossTech/ln-channel-closure-prediction}}.
\end{abstract}

\begin{IEEEkeywords}
Bitcoin, Lightning Network, payment channels, channel closure prediction, temporal graph, link classification
\end{IEEEkeywords}

\section{Introduction}
The Lightning Network (LN) \cite{poon_dryja_2016} is a second-layer protocol on top of Bitcoin that moves most payments off-chain. Two users open a \emph{payment channel} by jointly locking Bitcoin on-chain, route an arbitrary number of off-chain payments through it, and eventually settle back on-chain by closing the channel.

\begin{figure*}[t]
    \centering
    \includegraphics{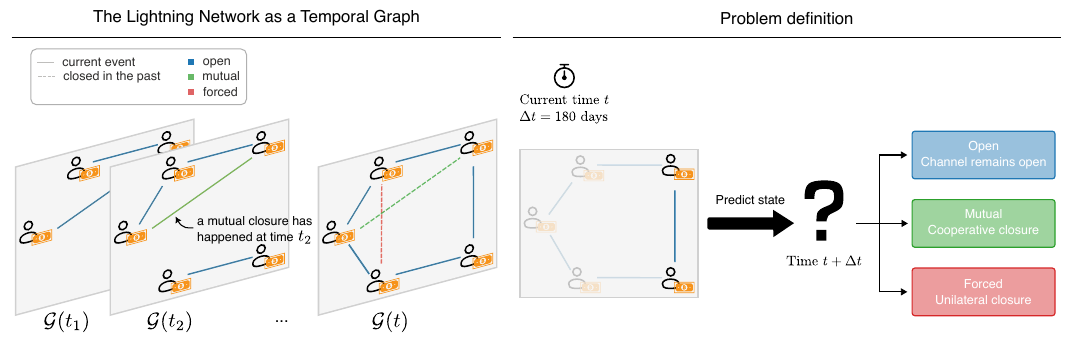}
    \caption{Overview of the channel closure prediction task. \emph{Left}: the Lightning Network evolves over time as channels open and close, forming a temporal graph. \emph{Right}: given the current graph state at time $t$, we predict whether each open channel will remain open, close cooperatively (mutual), or be force-closed within a window $\Delta t$.}\label{fig:teaser}
\end{figure*}

A \emph{mutual closure} settles the channel cooperatively and releases the funds immediately, while a \emph{forced closure} is initiated unilaterally, typically because one party is unresponsive or a dispute arises, and locks the initiating party's funds for a timelock period of days to weeks. Forced closures are costly: they consume on-chain fees, freeze liquidity that could otherwise be routed, and temporarily reduce network capacity. Anticipating them is therefore of practical interest for node operators, routing algorithms, and liquidity tooling.

The LN's topology and channel metadata are partially observable through its \emph{gossip protocol}, which broadcasts channel openings, closures, and periodic updates including fee policies, capacity, and disabled flags. This public information forms a temporal graph (\autoref{fig:teaser}), and raises the question: \emph{can we predict, from gossip data alone, whether a channel will remain open, close mutually, or be force-closed?}

Prior work has studied the LN's topology \cite{10.1145/3427796.3427837}, liquidity dynamics \cite{10.1145/3321705.3329812}, and applied Graph Neural Networks (GNNs) to snapshot-based LN tasks \cite{feichtinger2024benchmarkinggnnsusinglightning}, but none has explicitly modeled the temporal evolution of channel closures. We formalize the question above as a \emph{temporal link classification} task and conduct a systematic study of its predictability, benchmarking random baselines, gradient-boosted trees, an MLP, GNNs (static and temporal), and spectral graph encodings on a dataset of over two years of daily LN gossip snapshots that we release publicly. Our findings are: (i) the dominant predictive signals are temporal and behavioural, namely endpoint activity recency and per-node closure history, while static channel metadata is far less informative; (ii) graph topology, whether via message passing or spectral encodings, does not improve over a simple MLP using per-channel and per-node features; and (iii) overall performance remains moderate, reflecting a fundamental information gap: the signals most relevant to closure decisions (balances, payment failures, node uptime) are private by design.

\section{Problem statement}
We consider the definition of a \emph{temporal graph} as defined in \cite{DBLP:journals/corr/abs-2006-10637}, namely a set of events occurring at various timestamps that together build the final graph structure:
\begin{equation*}
    \gG = \{ x(t_m) : t_{m-1} \leq t_m \leq t_{m+1}, \text{ for } m \in [1, 2, \dots] \}
\end{equation*}
Each event $x(t)$ belongs to one of two types: a \emph{node-wise event} $\vv_i(t)$, involving the addition, deletion, or feature update of a node; or an \emph{interaction event} $\ve_{ij}(t)$, representing the addition or removal of an edge (i.e., a payment channel) between two nodes $i$ and $j$. A graph at time $t$, denoted $\gG(t)$, is defined by the pair $\left( \gV(t), \gE(t) \right)$, where $\gV(t) = \{ i : \vv_i(t_m) \in \gG \text{ and } t_m \leq t \}$ is the set of nodes present up to time $t$, and $\gE(t) = \{ (i, j) : \ve_{ij}(t_m) \in \gG \text{ and } t_m \leq t \}$ is the set of directed edges up to time $t$. Since payments in the LN flow in both directions, if $\ve_{ij}(t_m)$ represents an edge from $i$ to $j$, there also exists $\ve_{ji}(t_m)$ in the opposite direction. We denote a channel opening at time $t_m$ as $\ve_{ij}^+(t_m)$ and a channel closure as $\ve_{ij}^-(t_m)$.

At the \emph{current time} $t$, a model forecasts some property at the future \emph{query time} $t + \Delta_t$, where $\Delta_t$ is a configurable lookahead window. The task at hand can be formulated as a \emph{temporal link classification} problem: for each edge that is \emph{open} at the current timestamp $t_m$, the objective is to predict its state as \texttt{open}, \texttt{mutual}, or \texttt{forced} (see \autoref{sec:classes}) at the query time $t_m + \Delta_t$. In our experiments we set $\Delta_t = 180$ days, a choice we motivate and ablate in \autoref{sec:ablation}. An edge is \emph{open} at timestamp $t_m$ if its most recent interaction event up to $t_m$ is a channel opening. Formally, let $\gE^{+}(t_m) = \{(i,j)\mid\exists\,\ve_{ij}^{+}(t)\text{ with }t\le t_m\}$ and $\gE^{-}(t_m) = \{(i,j)\mid\exists\,\ve_{ij}^{-}(t)\text{ with }t\le t_m\}$. The set of open edges is then $\gE_{\text{open}}(t_m)=\gE^{+}(t_m)\setminus\gE^{-}(t_m)$. An \emph{open} edge is thus a persistent state of a channel, whereas interaction events are single occurrences that modify $\gG(t_m)$.

\section{Dataset}\label{sec:dataset}

We collect daily snapshots\footnote{On some days, data were collected twice at different times.} of the LN from its gossip messages, covering the period from June 9, 2022, to October 14, 2024. The raw data comprises \num{693277} directed events (channel openings and closures) recorded across \num{874} timestamps, involving \num{36170} unique nodes.

\textbf{Handling the initial snapshot.}\quad%
The first gossip snapshot (June 9, 2022) captures the accumulated history of the LN at that point, containing \num{358994} events (over half the dataset) at a single timestamp. To handle this, we adopt a \emph{warm-start} strategy: first-day events initialize the graph state (populating the set of open channels and node-level statistics) but are excluded from training, validation, and testing. This way, models learn from genuinely temporal activity while retaining the network's structure at the start of the observation period.

\textbf{Parallel channels.}\quad%
The LN is naturally a multigraph, where two nodes can maintain multiple channels simultaneously. To reduce it to a simple graph, we remove all node pairs that have more than one channel between them. This affects approximately 3\% of node pairs and reduces the total event count by roughly 20\%, but is necessary as the GNN architectures we benchmark do not support parallel edges, and it removes the ambiguity of which channel's properties to use for a node pair. We discuss the implications of this simplification in \autoref{sec:conclusion}.

\textbf{Chronological split.}\quad%
To prevent information leakage, we split the remaining data chronologically into training, validation, and test sets using a $70\%/15\%/15\%$ partition of the timeline. Each event's \texttt{channel\_status} attribute (opening vs.\ closure) is derived from gossip and available at event time, so using it leaks no future information.

\textbf{Labeling.}\quad%
We assign labels to open edges based on their future status. For any edge that is open at time $t$, we check whether a closing event involving that same edge occurs within the next $\Delta_t = 180$ days. If a closure is found, the edge is labeled according to the corresponding closure type (\texttt{forced} or \texttt{mutual}); otherwise, it remains labeled as \texttt{open}.

\subsection{Classes}\label{sec:classes}
\begin{figure}
    \centering
    \includegraphics{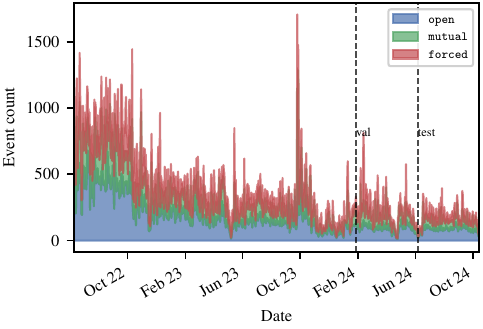}
  \caption{Distribution of label counts over time. The plot shows the temporal evolution of the three labels (\texttt{forced}, \texttt{mutual}, \texttt{open}) in the dataset, with vertical dashed lines indicating the starting points of the validation and test periods.}\label{fig:labels_time}
\end{figure}

Each open channel is assigned one of the following classes, reflecting its state in the LN:
\begin{itemize}
    \item \texttt{open}: The channel is operational within $\Delta_t$.
    \item \texttt{mutual}: Mutual closure agreement within $\Delta_t$.
    \item \texttt{forced}: The channel is unilaterally closed within $\Delta_t$.
\end{itemize}
A small fraction ($< 0.01\%$) of closures are classified as \texttt{penalty} closures, which we merge into the \texttt{forced} class.

\begin{figure*}
  \centering
  \includegraphics{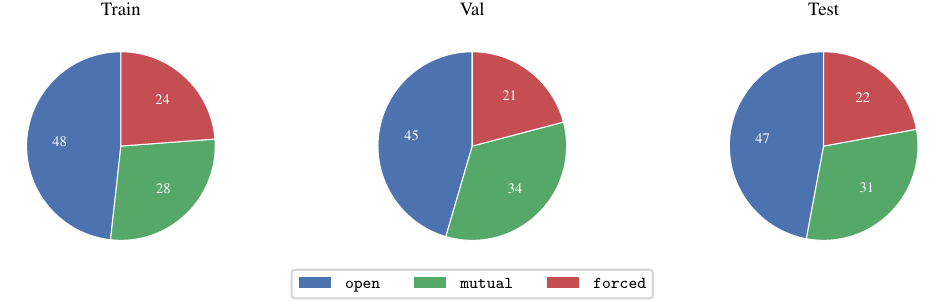}
  \caption{Daily average distribution of the three classes (\texttt{open}, \texttt{forced}, \texttt{mutual}) for the train, validation, and test splits. Proportions are computed by first considering daily fractions, then averaging across all days within each split.}\label{fig:labels_per_split}
\end{figure*}

\autoref{fig:labels_time} shows the distribution of event labels over time. Although the total number of events decreases as time progresses, the relative proportions remain fairly consistent, both over time and across the three temporal splits (\autoref{fig:labels_per_split}). Excluding the initial snapshot, \texttt{open} channels account for approximately 47\%, \texttt{mutual} closures for 30\%, and \texttt{forced} closures for 23\% of events. However, at prediction time, when the model must classify \emph{all currently open edges}, the distribution is heavily skewed: roughly $83\%$ of open edges remain \texttt{open}, with about $9\%$ eventually closing as \texttt{mutual} and $8\%$ as \texttt{forced}. We ablate strategies for handling this imbalance during training in \autoref{sec:ablation} (\autoref{tab:imbalance}).

\subsection{Node and edge features}\label{sec:features}
Each event carries features at both the channel (edge) and endpoint (node) levels, as reported in the corresponding gossip message. Per channel, we keep the on-chain timestamps (\texttt{ts}, \texttt{height}), the locked \texttt{capacity}, and the funding block's \texttt{block\_avg\_fee\_rate}; identifiers and labels (\texttt{transaction\_id/vout}, \texttt{channel\_status}, \texttt{event\_label}, \texttt{gossip\_ts}) are kept as metadata for bookkeeping and as the prediction target, but are not given to the model. Per endpoint we keep, separately for source and destination, the routing-policy parameters declared in gossip: base and proportional fees (\texttt{fee\_base\_msat}, \texttt{fee\_rate\_milli\_msat}), HTLC bounds (\texttt{min\_htlc}, \texttt{max\_htlc\_msat}), \texttt{time\_lock\_delta}, the \texttt{disabled} flag, the timestamp of the latest gossip update for that direction (\texttt{last\_update}), and the node's LN \texttt{implementation}. We preserve channel directionality and thus treat (\texttt{src}, \texttt{dst}) separately from (\texttt{dst}, \texttt{src}). All features are taken directly from gossip, and the dataset is sorted by \texttt{gossip\_ts} to maintain chronological order.

\section{Methodology}\label{sec:methodology}
Unlike most temporal graph benchmarks \cite{huang2023temporal, gastinger2024tgb}, which focus on \emph{link prediction}, i.e., whether an edge will form between two nodes, our task assumes the channel exists and classifies its future state (\texttt{open}, \texttt{mutual}, or \texttt{forced}). Class imbalance is a key challenge, as most channels remain open at any given time, with relatively few closing within a prediction window.

Formally, given a model $M$, a temporal window $\Delta_t$, the current timestamp $t$, and the graph $\gG(t)$ of open channels observed up to time $t$, the task is defined as follows:
\begin{tcolorbox}
For each \textit{open} edge $e \in \gE_{\text{open}}(t)$, let $t^\prime = t + \Delta_t$ be the query time:
\begin{enumerate}[label=\arabic*.]
    \item compute model prediction $s = M(e, t^\prime)$ and apply a softmax over the possible classes,
    \item evaluate $s$ using classification metrics (e.g., F1-score, accuracy).
\end{enumerate}
\end{tcolorbox}
In other words, for each open channel at time $t$, we want to predict its status at $t' = t + \Delta_t$ based on all events observed up to $t$. At each training step, the model predicts the status of \emph{all} currently open edges, not just those involved in the current batch's events. This differs from standard link prediction, which typically scores only a sampled subset of edges per step, and closely reflects deployed use.

We build our temporal evaluation pipeline on the Temporal Graph Network (TGN) framework \cite{DBLP:journals/corr/abs-2006-10637}, adapting two of its components for our setting. First, we replace the original \emph{neighbor loader}, which samples a fixed number of recent neighbors for queried nodes, with a variant that maintains the full set of currently open edges, inserting channels as they open and removing them as they close. At each step, this loader provides all open edges for prediction. Second, we replace TGN's learned RNN-based memory module with a simpler, non-parametric \emph{feature storage} that accumulates event counts (\texttt{open}, \texttt{forced}, \texttt{mutual}) per node as channels are opened and closed over time, providing a lightweight temporal summary of each node's history. All learned models share this temporal infrastructure and differ only in how they produce predictions from the current graph state.

\subsection{Baselines}

\textbf{Random baselines.}\quad%
We consider three non-learned baselines:
\begin{enumerate*}[label=\roman*)]
    \item \texttt{uniform}, sampling labels uniformly at random;
    \item \texttt{stratified}, sampling labels based on observed class frequencies in the training set;
    \item \texttt{majority}, always predicting the most frequent class (\texttt{open}).
\end{enumerate*}

\subsection{MLP predictor}

Our primary model is a multi-layer perceptron (MLP) that classifies each open edge independently, without any graph-based message passing. For each open edge $(i,j)$ at time $t$, the input feature vector is the concatenation of:
\begin{itemize}
    \item \emph{Edge features}: channel properties from the gossip protocol, including capacity, fee policies (base fee, fee rate), disabled flags, timelocks, min/max HTLC values, etc.
    \item \emph{Node features}: for each endpoint $i$ and $j$, the running counts of events of each type (\texttt{count\_open}, \texttt{count\_forced}, \texttt{count\_mutual}) accumulated from the feature storage. These summarise each node's opening and closure history up to time $t$ (3 dimensions per node).
    \item \emph{Temporal encodings}: the channel age $t - t_{\text{open}}$ (\texttt{edge\_age}), and the source and destination \emph{recency} $t - t_{\text{last\_update},i}$ (\texttt{src\_recency}) and $t - t_{\text{last\_update},j}$ (\texttt{dst\_recency}), where $t_{\text{last\_update},k}$ is the timestamp of the most recent gossip event involving node $k$. Each of these three scalars is passed through a learnable time encoder ($3 \times d_{\text{time}}$ dimensions in total).
\end{itemize}

\subsection{Gradient-boosted trees}

As tabular baselines we also include two gradient-boosted decision tree classifiers, XGBoost \cite{10.1145/2939672.2939785} and LightGBM \cite{NIPS2017_6449f44a}, both receiving exactly the same input vector as the MLP. We replay the training events to populate the neighbor loader and feature storage, then fit each model once on the snapshot of all currently open edges at the end of training, using the oracle's closure labels. At test time the model predicts at each timestamp from the features extracted from the current snapshot. Both models use $500$ trees of depth $6$, learning rate $0.1$, and the same per-class loss weights $[1,5,5]$ as the neural baselines.\looseness-1

\subsection{Graph-based models}

We evaluate two GNN variants and a spectral baseline, covering different ways of injecting structure into the prediction.

\textbf{Static GNN.}\quad%
A GraphSAGE \cite{NIPS2017_5dd9db5e} network operating on the current graph of open edges, with node embeddings initialised from current degrees and aggregated through the GraphSAGE layers before being fed to a prediction MLP. It does not use edge features or temporal encodings, isolating the predictive value of graph structure.

\textbf{TGN.}\quad%
The TGN uses the same input features as the MLP and additionally computes node embeddings via attention-based message passing over the current graph of open edges, using edge features as attention inputs. These embeddings are concatenated with the edge features, node features, and temporal encodings in the prediction MLP, letting the model capture structural patterns inaccessible to the edge-level MLP.

\textbf{Spectral encodings.}\quad%
As an alternative to message passing, we augment the MLP with the top-$k$ eigenvectors of the normalised Laplacian of the open-edges graph, concatenating each endpoint's spectral position $[\phi_i, \phi_j] \in \mathbb{R}^{2k}$ to the input. These encodings capture each node's structural role in the topology without iterative aggregation. We use $k=16$, recomputed periodically as the graph evolves.

\section{Experimental setup and results}\label{sec:setup_results}
We assess model performance using the macro-average F1-score, which is well suited for handling class imbalance. Edge features are preprocessed with a log-transform followed by min-max scaling fitted on the training set. All learned models are trained for 30 epochs with a weighted cross-entropy loss (weights $[1, 5, 5]$), optimized via Adam (lr $= 10^{-4}$, weight decay $= 10^{-5}$) with linear warmup over 1000 steps. We use hidden dimension 128 and temporal encoding dimension 128. We report means and standard deviations over 3 seeds. All models train on an 8-core CPU (no GPU): ${\sim}1.2$\,h for the MLP vs ${\sim}4$--$12.5$\,h for the graph-based models and ${<}1$\,min for the boosted trees.

\begin{table}[t]
\caption{Performance on the test set: per-class and macro-average F1-scores as mean $\pm$ std over 3 seeds.}\label{tab:results}
\centering
\footnotesize
\setlength{\tabcolsep}{4pt}
\begin{tabular}{@{}lcccc@{}}
\toprule
\textbf{Model} & \texttt{open} & \texttt{forced} & \texttt{mutual} & \textbf{Macro F1} \\
\midrule
Uniform & $0.48_{\pm.0003}$ & $0.13_{\pm.0003}$ & $0.14_{\pm.0008}$ & $0.25_{\pm.0004}$ \\
Majority & $0.91_{\pm.0000}$ & $0.00_{\pm.0000}$ & $0.00_{\pm.0000}$ & $0.30_{\pm.0000}$ \\
Stratified & $0.76_{\pm.0003}$ & $0.10_{\pm.0005}$ & $0.12_{\pm.0005}$ & $0.32_{\pm.0001}$ \\
\midrule
Static GNN & $0.84_{\pm.0086}$ & $0.14_{\pm.0057}$ & $0.06_{\pm.0029}$ & $0.35_{\pm.0002}$ \\
TGN & $0.73_{\pm.0234}$ & $0.15_{\pm.0033}$ & $0.22_{\pm.0012}$ & $0.36_{\pm.0071}$ \\
MLP + Spectral & $0.69_{\pm.0200}$ & $0.15_{\pm.0027}$ & $0.21_{\pm.0009}$ & $0.35_{\pm.0059}$ \\
XGBoost & $0.87_{\pm.0000}$ & $0.12_{\pm.0000}$ & $0.10_{\pm.0000}$ & $0.36_{\pm.0000}$ \\
LightGBM & $0.86_{\pm.0000}$ & $0.11_{\pm.0000}$ & $0.12_{\pm.0000}$ & $0.36_{\pm.0000}$ \\
\textbf{MLP} & $0.80_{\pm.0042}$ & $0.13_{\pm.0029}$ & $0.20_{\pm.0018}$ & $\mathbf{0.38}_{\pm.0012}$ \\
\bottomrule
\end{tabular}
\end{table}

\begin{figure*}[t]
    \centering
    \includegraphics{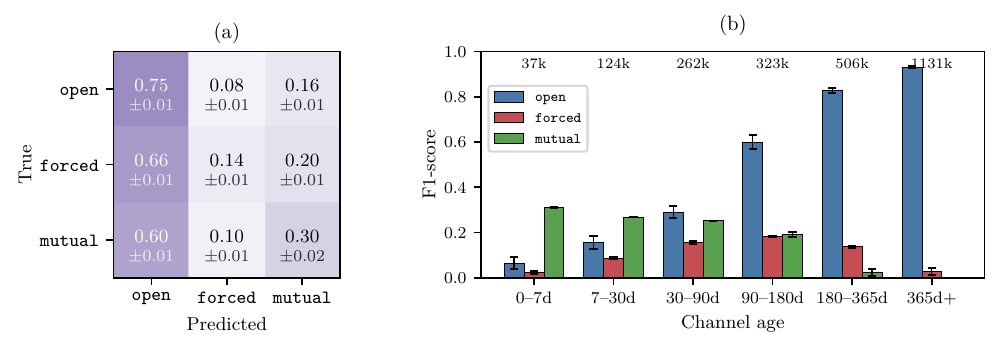}
    \caption{(a) Normalized confusion matrix for the MLP (\texttt{open}, \texttt{forced}, \texttt{mutual}). The model achieves high recall on \texttt{open} but struggles to distinguish closure types. (b) Per-class F1 binned by channel age at query time. Open F1 increases sharply with channel age, showing that long-lived channels are reliably predicted to remain open. Conversely, forced and mutual F1 decrease with age and approach zero in the oldest bin, where closures are rare.}\label{fig:analysis}
\end{figure*}

\subsection{Main results}

\autoref{tab:results} compares all models on the test set. For the MLP and TGN we report the best configuration from our layer ablation (\autoref{fig:layers}), while the other learned models use default hyperparameters. The MLP predictor achieves the best macro-average F1 of $0.38 \pm 0.001$. The TGN, which additionally computes node embeddings via GNN message passing, reaches $0.36 \pm 0.007$, still below the MLP. The static GNN and the MLP augmented with spectral positional encodings both achieve $0.35$. All learned models improve over the stratified random baseline ($0.32$), but the margin remains modest.

Notably, the MLP uses no graph information whatsoever, yet outperforms all graph-aware models and the gradient-boosted tree baselines that share its input features. The TGN's GNN message passing does not improve over the MLP despite access to neighborhood structure, and neither spectral positional encodings nor the static GNN help. Paired $t$-tests across seeds support this picture: the MLP's advantage is significant over the static GNN, the spectral variant, both gradient-boosted trees, and all random baselines (all $p < 0.05$), while the TGN remains statistically indistinguishable from the MLP ($\Delta = 0.012$ macro F1, $95\%$ CI $[-0.008, 0.032]$, $p = 0.12$, lower in every seed), confirming that message passing brings no measurable benefit. Graph topology thus provides little additional signal beyond per-channel and per-node features, as we investigate in the following ablations.

\autoref{fig:analysis}(a) shows the confusion matrix for the MLP. The model correctly identifies most \texttt{open} channels but frequently confuses \texttt{forced} and \texttt{mutual} closures with each other and with \texttt{open}, suggesting that the gossip features do not clearly distinguish the two closure types. \autoref{fig:analysis}(b) breaks down per-class F1 by channel age. The \texttt{open} F1 increases sharply with age, reaching $0.93$ for channels older than a year, as the model learns that long-lived channels rarely close. Conversely, \texttt{mutual} F1 is highest for recently-opened channels and declines steadily with age, while \texttt{forced} F1 peaks for medium-aged channels (90--180 days); both drop to near zero for the oldest bin, where closures are rare and difficult to detect.

\subsection{Ablation studies}\label{sec:ablation}

We now investigate which factors drive performance, through four complementary ablations: feature groups, model depth, prediction window, and class imbalance handling.

\textbf{Feature groups.}\quad%
To understand which components drive the MLP's performance and why graph-based models underperform, we conduct a feature ablation study. All configurations share the same temporal pipeline (neighbor loader and feature storage) and the same prediction MLP, differing only in which feature groups are exposed to the prediction head and whether GNN message passing is enabled on top.

\begin{table}[t]
\caption{Feature ablation using the best MLP and TGN configurations. We progressively add feature groups and assess whether GNN message passing adds value on top of them. Mean $\pm$ std over 3 seeds.}\label{tab:ablation}
\centering
\footnotesize
\setlength{\tabcolsep}{4pt}
\begin{tabular}{@{}lccccc@{}}
\toprule
\textbf{Features} & \textbf{GNN} & \textbf{Edge} & \textbf{Time} & \textbf{Node} & \textbf{Macro F1} \\
\midrule
Time only & -- & -- & \checkmark & -- & $0.36_{\pm.0005}$ \\
+ Edge feat. & -- & \checkmark & \checkmark & -- & $0.36_{\pm.0017}$ \\
\textbf{+ Node feat.} & -- & \checkmark & \checkmark & \checkmark & $\mathbf{0.38}_{\pm.0012}$ \\
\midrule
Edge + Time + GNN & \checkmark & \checkmark & \checkmark & -- & $0.37_{\pm.0023}$ \\
All + GNN & \checkmark & \checkmark & \checkmark & \checkmark & $0.36_{\pm.0071}$ \\
\bottomrule
\end{tabular}
\end{table}

The top half of \autoref{tab:ablation} progressively adds feature groups to the MLP. Starting from the time-only baseline, edge features alone do not help; the largest gain comes from adding the per-node event counts, which lift performance to $0.38$ and reach the full MLP. The accumulated history of how each endpoint has behaved is thus the dominant signal, with edge features contributing only in combination with it.

The bottom half enables GNN message passing on top of these same features, yielding the TGN architecture. With only edge and time features, GNN message passing slightly helps (Edge + Time + GNN reaches $0.37$, above $0.36$ for the same features without the GNN), but once node-level event counts are added, the GNN no longer brings any benefit and performance drops back to $0.36$. The MLP augmented with spectral encodings (\autoref{tab:results}) similarly fails to improve over the baseline MLP. Overall, graph aggregation can compensate when richer node features are missing, but does not unlock performance beyond what per-node history and edge features already provide.

\begin{figure}[t]
    \centering
    \includegraphics{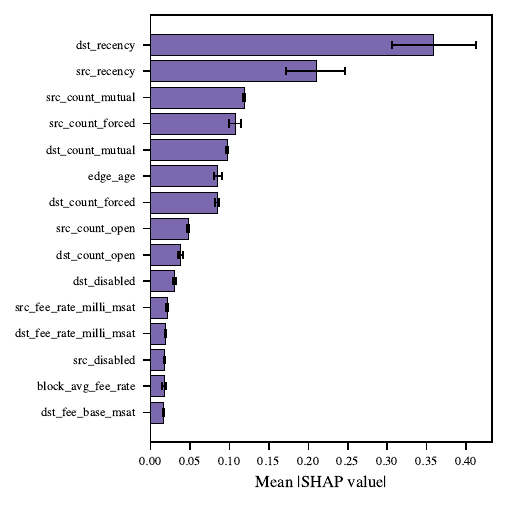}
    \caption{Feature importances for the trained MLP, computed as the mean absolute SHAP value over test edges, averaged over three seeds (error bars: standard deviation). Behavioural and temporal signals (node recency, per-node closure counts, channel age) dominate over static channel metadata.}\label{fig:feature_importance}
\end{figure}

To complement the feature-group ablation, we also estimate the importance of \emph{individual} input features for the trained MLP. We compute SHAP values via gradient-based attribution on a sample of currently-open edges at the end of training, and report the mean absolute SHAP value per feature. \autoref{fig:feature_importance} shows the top fifteen features. The dominant signals are how recently each endpoint was active (\texttt{src\_recency}, \texttt{dst\_recency}) and the per-node history of past closures (\texttt{src/dst\_count\_mutual}, \texttt{src/dst\_count\_forced}), together with the channel's age. Static channel metadata such as fee policies, capacity, disabled flags, and timelocks appear well below the top of the ranking. The ranking is highly stable across seeds (pairwise Spearman rank correlation $\rho \geq 0.99$, with an identical top five).

\begin{figure}[t]
    \centering
    \includegraphics{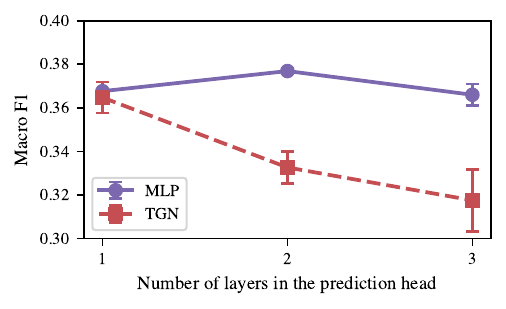}
    \caption{Effect of the prediction head depth, where $1$ corresponds to a single linear layer (logistic regression). For the TGN, we fix the GNN depth to 1 and vary the prediction MLP head. Deeper heads perform worse in both cases. Error bars show the standard deviation over three seeds.}\label{fig:layers}
\end{figure}

\textbf{Model depth.}\quad%
We also vary the depth of the prediction head, ranging from a single linear layer (logistic regression) to three layers, for both the MLP and the TGN (\autoref{fig:layers}). For the TGN, we fix the GNN depth to its best value ($1$ layer) and vary only the head. A shallow MLP with one hidden layer achieves the best performance ($0.38$), only marginally above plain logistic regression ($0.37$), while deeper architectures degrade. The TGN follows a similar trend, peaking with a linear head ($0.36$) and degrading with depth. At every setting the MLP outperforms the TGN, confirming that neither additional model capacity nor GNN message passing helps on this task.

\begin{figure}[t]
    \centering
    \includegraphics{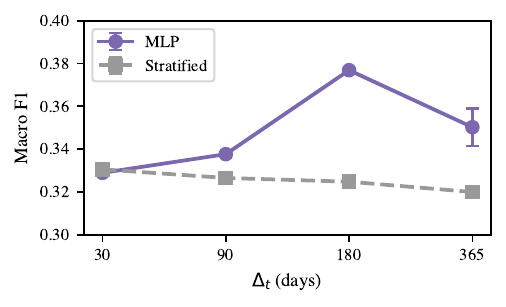}
    \caption{Effect of the prediction window $\Delta_t$ on the MLP compared to the stratified baseline. The MLP matches the baseline at $\Delta_t = 30$ days and outperforms it at all longer horizons, with the largest gap at $\Delta_t = 180$ days. Error bars show the standard deviation over three seeds.}\label{fig:delta_t}
\end{figure}

\textbf{Prediction window.}\quad%
We also vary the prediction window $\Delta_t \in \{30, 90, 180, 365\}$ days (\autoref{fig:delta_t}). The MLP matches the stratified baseline at $\Delta_t = 30$ days and outperforms it at all longer horizons, with the largest gap at $\Delta_t = 180$ days. At short horizons few channels close, leaving little signal to learn; at $\Delta_t = 365$ days the uncertainty over a full year dilutes the predictive value of current gossip features. The 180-day window is also comfortably above the typical closure timescale in our dataset: among channels that eventually close, the median lifetime is $73$ days and roughly $76\%$ close within $180$ days, so the window captures the bulk of closure events. For queries in the final $\Delta_t$ of the timeline, part of the label window extends beyond the end of data collection: if a channel closes in that unobserved period, the closure is invisible to us and the edge is labelled \texttt{open}. This truncation never affects training labels, since the training period ends more than $\Delta_t$ before the data does, while late validation and test queries observe progressively shorter windows. We therefore adopt $\Delta_t = 180$ days for the main experiments, as a window that is both informative and well aligned with the natural closure timescale of the network.

\begin{table}[t]
\caption{MLP ablation over strategies for handling class imbalance. Class weights are the coefficients in the weighted cross-entropy loss; downsampling keeps $r$ \texttt{open} edges per closing edge in the loss. Mean $\pm$ std over 3 seeds.}\label{tab:imbalance}
\centering
\footnotesize
\begin{tabular*}{\columnwidth}{@{\extracolsep{\fill}}llcc@{}}
\toprule
\textbf{Strategy} & \textbf{Weights} & \textbf{Downsample} & \textbf{Macro F1} \\
\midrule
Unweighted & $[1, 1, 1]$ & -- & $0.30_{\pm.0000}$ \\
Mild asymmetric & $[1, 2, 3]$ & -- & $0.33_{\pm.0036}$ \\
Frequency-aware & $[1, 3, 6]$ & -- & $0.34_{\pm.0026}$ \\
\textbf{Moderate (default)} & $\mathbf{[1, 5, 5]}$ & -- & $\mathbf{0.38}_{\pm.0012}$ \\
Extreme symmetric & $[1, 10, 10]$ & -- & $0.34_{\pm.0061}$ \\
Inverse frequency & $[1, 8.5, 17]$ & -- & $0.25_{\pm.0073}$ \\
\midrule
Moderate & $[1, 5, 5]$ & $r=5$ & $0.38_{\pm.0010}$ \\
Balanced & $[1, 1, 1]$ & $r=1$ & $0.37_{\pm.0050}$ \\
\bottomrule
\end{tabular*}
\end{table}

\textbf{Class imbalance.}\quad%
Finally, we ablate different strategies for handling the severe class imbalance (\autoref{tab:imbalance}). Without any class weighting, the model collapses to predicting \texttt{open} for all edges (macro F1 $= 0.30$, the majority baseline). Mild reweighting ($[1, 2, 3]$ or $[1, 3, 6]$) only partially compensates and stays close to the stratified baseline. Symmetric moderate weights $[1, 5, 5]$ are the sweet spot, while more aggressive weighting hurts: $[1, 10, 10]$ falls below the moderate setting, and inverse-frequency weights $[1, 8.5, 17]$ collapse below the majority baseline. Balanced downsampling ($r=1$, uniform weights) recovers most of the performance ($0.37$), and combining it with the default weights matches the weighted loss. Overall, weighted cross-entropy with $[1, 5, 5]$ is the simplest and most effective choice.

\subsection{Discussion}

The consistent finding across all experiments is that temporal and behavioural node-level signals are the only features that meaningfully drive closure prediction, which has a natural interpretation in the LN's design.

The gossip protocol broadcasts channel-level metadata (fee policies, capacity, disabled flags, timelocks) but reveals almost nothing about the \emph{activity} flowing through channels. Channel balances, payment volumes, routing failures, and node uptime are all private. Yet these are precisely the factors most likely to trigger a forced closure: a node going offline, a payment dispute, or a depleted channel balance. Since this information is invisible in the gossip data, the model falls back on the closest observable proxies, namely how often each endpoint has been involved in past closures and how recently it has been active. The static channel parameters (fees, capacity, timelocks), by contrast, are largely set at opening time and rarely revisited, and our SHAP analysis confirms that the model relies on them only marginally.

The same logic explains why graph topology adds little. In networks where neighbourhoods are informative, such as social networks, GNNs can leverage structure. In the LN, however, a node's neighbours reveal little about whether \emph{this specific channel} will be force-closed, because the determining information stays private to the two endpoints and is not propagated through gossip. The moderate overall performance (best macro F1 of $0.38$) is therefore better understood as a \emph{fundamental information gap} than as a failure of the models. Substantially improving beyond this level would likely require access to private node-side data, such as channel balances, payment histories, or per-node uptime, that the gossip protocol is intentionally designed not to expose. In practice, a model at this level is therefore best used to \emph{rank} channels by closure risk, e.g.\ to prioritise monitoring and rebalancing, rather than as a per-channel classifier.

\section{Related work}
\textbf{Machine learning for the Lightning Network.}\quad
Since its inception, the LN has been the subject of extensive research, focusing on both its topology \cite{10.1145/3427796.3427837, 10.1007/978-3-030-37110-4_1} and liquidity dynamics \cite{10.1145/3321705.3329812}. ML techniques have increasingly been applied to LN-specific problems: \cite{rossi2024channel} explore different methods to predict channel balances, \cite{salahshour2024jointcombinatorialnodeselection} employ reinforcement learning for joint node selection and resource allocation, and \cite{singh2024bayesian} leverage probabilistic modeling to optimize payment probing. Graph-based ML methods have also been applied to the LN: \cite{feichtinger2024benchmarkinggnnsusinglightning} benchmark GNNs on LN-specific tasks using snapshot-based datasets. However, prior studies have not explicitly incorporated the temporal dimension of the LN nor addressed the specific problem of predicting channel closure types. Our work fills this gap by modeling the LN as a continuous-time dynamic graph and providing a systematic study of closure predictability using publicly available gossip data.

\textbf{Temporal graph neural networks.}\quad
Early approaches to temporal graphs modeled them as sequences of snapshots and applied GNNs to discrete representations \cite{Pareja_Domeniconi_Chen_Ma_Suzumura_Kanezashi_Kaler_Schardl_Leiserson_2020, 10.1007/s10489-021-02518-9, 10.1145/3447548.3467422, 10.1145/3534678.3539300, 10.1145/3580305.3599551}, commonly referred to as Discrete-Time Dynamic Graphs (DTDGs). More recent approaches model temporal graphs continuously as event sequences, termed Continuous-Time Dynamic Graphs (CTDGs) \cite{trivedi2018dyrep, 10.1145/3292500.3330895, Xu2020Inductive, DBLP:journals/corr/abs-2006-10637, yu2023towards, cong2023do}. The Temporal Graph Network (TGN) \cite{DBLP:journals/corr/abs-2006-10637} provides a general framework combining memory modules, message passing, and temporal encoding, and has become widely adopted. Standardized benchmarks \cite{huang2023temporal, gastinger2024tgb} have facilitated progress, primarily on link existence prediction and node classification tasks. Our work applies temporal graph methods to a different task, link \emph{classification}, and provides evidence that, for this particular application, the temporal and edge-level components are more valuable than graph aggregation.

\section{Conclusion and Further Research}\label{sec:conclusion}

We studied predicting channel closure types in the Lightning Network from publicly available gossip data, formalising the task as a temporal link classification problem, constructing a dataset spanning two years of LN activity, and benchmarking approaches ranging from random baselines and gradient-boosted trees to static and temporal GNNs and spectral encodings.

Our experiments revealed that the dominant predictive signals are temporal and behavioural (endpoint activity recency, per-node history of past closures, channel age), while static channel metadata and graph topology contribute much less. The best-performing model is a simple MLP operating on edge- and node-level features without any graph message passing, reaching a macro-average F1 of $0.38$.

\textbf{Limitations.}\quad%
As discussed in \autoref{sec:setup_results}, the moderate overall performance reflects an information gap rather than a model limitation: the factors most likely to trigger forced closures (balance depletion, routing failures, node downtime) are private by design and not disclosed by gossip. Two scope restrictions follow. First, current GNN architectures do not handle parallel edges, forcing us to remove them (\autoref{sec:dataset}) and discard roughly 20\% of events on the ${\sim}3\%$ of node pairs with multiple channels. These pairs typically involve large, well-connected routing nodes, whose closure behaviour may differ from the single-channel pairs we study. Second, closure decisions also respond to conditions outside the gossip network: closing a channel requires an on-chain transaction, so mempool congestion and fee spikes directly change the cost, and hence the incentives, of closing. Beyond the funding block's average fee rate (ranked low by SHAP), our features do not capture these time-varying market conditions.

\textbf{Future directions.}\quad%
Beyond closure prediction, we believe temporal modelling of the channel graph is a valuable tool for operators, whose outcomes (payment reliability, earned routing fees) are functions of how the graph evolves over time. A straightforward next step is to incorporate time-varying on-chain signals (e.g.\ mempool congestion and fee-market conditions). Further directions include testing richer structural descriptors (e.g.\ centrality or community features), exploring node-local prediction using private data such as local balance histories or payment failure logs that are available only to the channel endpoints, and studying how closure patterns evolve as the LN's topology and usage shift over time.

\bibliographystyle{IEEEtran}
\bibliography{bibliography}

\end{document}